\documentclass[runningheads]{llncs}

\usepackage{eccv}

\usepackage{eccvabbrv}
\usepackage{graphicx}
\usepackage{booktabs}
\usepackage{algorithm}
\usepackage[noend]{algpseudocode}
\usepackage{amsmath} % For general math features
\usepackage{amssymb} % For \mathbb and other symbols
\usepackage{bm}      % For bold math
\usepackage[accsupp]{axessibility}

\usepackage[breaklinks=true,colorlinks=false,pdfborder={0 0 1}]{hyperref}
\hypersetup{
    linkcolor=red,          % Assuming you want internal links (e.g., figures) in red
    citecolor=green,        % Color of links to bibliography (paper citations) in green
    filecolor=magenta,      % Color of file links
    urlcolor=cyan,          % Color of external links
    linkbordercolor={1 0 0},  % Red border color for links
    citebordercolor={0 1 0},  % Green border color for citations
    urlbordercolor={0 1 1}    % Cyan border color for URLs
}

\begin{document}

\title{AnimateMe: 4D Facial Expressions via Diffusion Models}
\titlerunning{AnimateMe: 4D Facial Expressions via Diffusion Models}
\author{Dimitrios Gerogiannis \and
Foivos Paraperas Papantoniou \and
Rolandos Alexandros Potamias \and
Alexandros Lattas \and
Stylianos Moschoglou \and
Stylianos Ploumpis \and 
Stefanos Zafeiriou
}
\authorrunning{D. Gerogiannis et al.}
\institute{Imperial College London, UK\\
\email{\{d.gerogiannis22, f.paraperas, r.potamias,  a.lattas, s.moschoglou, s.ploumpis, s.zafeiriou\}@imperial.ac.uk}}

\maketitle

\begin{abstract}
    The field of photorealistic 3D avatar reconstruction and generation has garnered significant attention in recent years; however, animating such avatars remains challenging. Recent advances in diffusion models have notably enhanced the capabilities of generative models in 2D animation. In this work, we directly utilize these models within the 3D domain to achieve controllable and high-fidelity 4D facial animation. By integrating the strengths of diffusion processes and geometric deep learning, we employ Graph Neural Networks (GNNs) as denoising diffusion models in a novel approach, formulating the diffusion process directly on the mesh space and enabling the generation of 3D facial expressions. This facilitates the generation of facial deformations through a mesh-diffusion-based model. Additionally, to ensure temporal coherence in our animations, we propose a consistent noise sampling method. Under a series of both quantitative and qualitative experiments, we showcase that the proposed method outperforms prior work in 4D expression synthesis by generating high-fidelity extreme expressions. Furthermore, we applied our method to textured 4D facial expression generation, implementing a straightforward extension that involves training on a large-scale textured 4D facial expression database.
\end{abstract}

\section{Introduction}
\label{sec:intro}

    In the field of computer graphics and human-computer interaction, 3D face modeling \cite{blanz2023morphable,blanz2003face, ploumpis2020towards,ploumpis2019combining, tran2018nonlinear,bouritsas2019neural} and animation are becoming increasingly crucial. With the evolution of digital interactions, there is a rising demand for realistic 3D avatars that can generate emotions with high fidelity. While controllable and dynamic 2D facial expression generation is well-studied \cite{f3agan,im2video,g3an,tulyakov2017mocogan,wgan2d,Bouzid_2022}, its 3D counterpart remains unexplored due to its inherent complexity. This lack of research emphasizes the need for advancements in 4D facial expression generation. While much of the 3D face animation domain focuses on speech animation \cite{cudeiro2019capture,karras2017audio,pham2017speech,tzirakis2020synthesising,Richard_2021_ICCV,peng2023emotalk,fan2022faceformer,Thambiraja_2023_ICCV,xing2023codetalker,stan2023facediffuser,park2023said,aneja2023facetalk,thambiraja20233diface,ma2024diffspeaker}, research on expression animation \cite{otberdout2022sparse,potamias2020learning} is limited. Although diffusion models have been sparingly applied to facial speech animation techniques \cite{stan2023facediffuser,park2023said,aneja2023facetalk,thambiraja20233diface,ma2024diffspeaker}, the results have shown considerable promise. In contrast, these models have been extensively and successfully employed in the field of 3D human motion, demonstrating their effectiveness and versatility \cite{tevet2022human,chen2023executing,azadi2023make,dabral2023mofusion,zhang2024motiondiffuse,shafir2023human}. Despite the proven success of diffusion models in 3D animation, their application in 4D facial expression remains unexplored. This gap in the literature inspired us to develop our method, aiming to investigate their potential in this domain. 

    However, typical diffusion model architectures are challenging to apply directly to 3D structures. Using meshes as our 3D face representation, we introduce the first 3D diffusion process tailored to operate directly within the mesh space, enabling the application of diffusion models for 4D facial expression generation. Our method achieves this by employing Graph Neural Networks (GNNs) as denoising diffusion models. This novel approach paves the way for broader utilization of GNNs in diffusion processes for mesh generation. Moreover, regarding temporal coherence in animations, our method diverges from the traditional diffusion models methods. Conventional video approaches \cite{ho2022video,ho2022imagen, singer2023makeavideo, zhou2023magicvideo, he2023latent, blattmann2023align} typically handle temporal dependencies via architectural adjustments incorporating temporal modules, generating frames collectively. In contrast, inspired by ideas presented in \cite{khachatryan2023text2video,wu2023latent,luo2023videofusion}, our approach modifies the traditional DDPM algorithm by introducing a straightforward yet effective sampling strategy tailored to our specific challenge, termed consistent noise sampling. This intuitive sampling strategy not only solidifies temporal coherence but also significantly improves the generation time.
    
    Our method capitalizes on diffusion models, presenting a marked divergence from the competition. Although some methods have been explored for 4D facial expression generation \cite{otberdout2022sparse,potamias2020learning}, they often fall short in producing high-fidelity extreme expressions, while our method successfully accomplishes it. This advantage is primarily attributed to the diffusion models' capability to handle complex distributions, such as the extreme deformations accompanying intense facial expressions. The strength of our model is further enhanced by utilizing the entirety of the mesh space. This offers superior capturing capabilities compared to traditional blendshapes \cite{cao_faceware} and landmarks \cite{otberdout2022sparse}. 
    
    In summary, our work offers the following key contributions:
    \begin{itemize}
        \item The first, to the best of our knowledge, diffusion process formulation operating directly on the mesh space with GNNs proposed as denoising models.
        \item The first, to the best of our knowledge, fully data-driven approach to customizable 4D facial expressions, utilizing diffusion models.
        \item A dynamic diffusion models sampling strategy for 3D facial animation, that is extended to both geometry and texture generation.
    \end{itemize}

\section{Related Work}
\label{sec:related}
    \subsection{3D Facial Animation Generation}
    \label{sec:related_animation}
    Since the introduction of the seminal 3DMMs \cite{blanz2003face}, multiple methods \cite{blanz2023morphable} have been proposed to model facial animation using expression blendshapes \cite{cao_faceware, Cheng_2018_CVPR}. Nonetheless, they train on static 3D meshes, and can only rely on unrealistic linear interpolation to represent 3D facial motion.

    Several methods have tried to tackle the limitations of 3D facial motion synthesis by utilizing either audio or speech features. In an early attempt to model speech features along with facial motion, the authors of \cite{karras2017audio} presented a subject-specific model to produce facial motion from audio but faced limitations in adapting to varied speakers. A similar method was proposed in \cite{cudeiro2019capture} where a static neutral mesh of a given identity was fed to the network for animation based on speech input features showcasing flexibility across diverse speakers. Several other approaches were built on top, that utilize RNNs along with facial action units \cite{pham2017speech} and LSTMs \cite{tzirakis2020synthesising}. Following studies significantly improved the realism of 3D speech animation by disentangling emotion from speech \cite{Richard_2021_ICCV,peng2023emotalk}, while Transformer-based approaches\cite{fan2022faceformer,Thambiraja_2023_ICCV,xing2023codetalker} have demonstrated promising results as well. Only very recently, a few works have focused on animating a 3D facial mesh via a diffusion process coupled with a voice input signal \cite{stan2023facediffuser,park2023said,aneja2023facetalk,thambiraja20233diface,ma2024diffspeaker}. Most of these works deal with denoising and generating animations based on input speech that is later mapped into blendshape expression parameters \cite{park2023said,stan2023facediffuser}. Only\cite{thambiraja20233diface} works directly on the mesh domain utilizing a motion decoder. Even though this line of work is promising for facial animation based on speech input, none of the aforementioned methods deal with labeled (guided) expression generation.

    Guided 4D expression generation is an understudied problem with only a few methods able to achieve satisfactory results. The work proposed in \cite{potamias2020learning} first attempted guided 4D expression generation on the 4DFAB dataset \cite{Cheng_2018_CVPR}, combining LSTMs with GNNs. It notably addressed the challenge of animating extreme expressions for the first time. While this work utilizes a GNN architecture as a decoder, similar to our method, our approach differentiates itself by adopting a mesh diffusion process, thereby enhancing the expression capturing capabilities. Similar to this research direction, the authors in \cite{otberdout2022sparse} proposed a two-stage framework employing a sparse-to-dense decoder mapping sparse 3D facial landmark displacements to dense ones for motion generation, complemented by a GAN architecture for landmark sequence creation driven by expression labels. Although effective, this method falls short of capturing extreme expressions and detailed facial features as efficiently as our approach, as demonstrated in our experiments section. This discrepancy is likely due to the superiority of diffusion models over GANs \cite{dhariwal2021diffusion}, and our comprehensive use of the entire mesh rather than just landmarks.
    
    \subsection{Diffusion Models for Animation}
    \label{sec:related_diffusion}
    Since the introduction of diffusion models in the 2D image domain \cite{ho2020denoising,song2020score}, the computer vision literature witnessed staggering advancements in image and video synthesis when compared to previous GAN-based approaches \cite{dhariwal2021diffusion,rombach2022high}. Regarding 2D animation, \cite{ho2022video} was the first work to extend diffusion models to video synthesis by adapting the traditional denoising architecture to accommodate 3D data. Following this pioneering work, many studies focused on maintaining temporal coherence in animation, while achieving high resolution and frame rate through a variety of methods. These include integrating temporal modules, refining denoising architectures for video data, implementing cascaded super-resolution techniques, extending text-to-image models to videos, extending the latent diffusion model paradigm to video diffusion models, or even combinations of these approaches, often leveraging the strengths of each to achieve superior outcomes \cite{ho2022imagen, singer2023makeavideo, zhou2023magicvideo, he2023latent, blattmann2023align}. The unique approaches of \cite{luo2023videofusion} and \cite{khachatryan2023text2video} diverge from the typical approaches to handling temporal coherence. The authors of \cite{luo2023videofusion} innovated by treating video frames as non-independent instances by refining the DDPM paradigm to introduce a shared base noise and a time-variant residual noise across frames. Similarly, \cite{khachatryan2023text2video} departs from traditional random sampling by enforcing motion dynamics between the latent codes. Together, these two approaches emphasize the benefits of using consistent noise patterns across frames serving as inspiration for our work.
    
    Extending diffusion models to generate animation in the 3D domain is an ambitious task given the vast amounts of 3D data required over time and the inherently greater complexity of 3D structures compared to images. Recent efforts have leveraged these models to generate motion for 3D models. Apart from the few diffusion-based 3D speech animation methods referenced in the previous subsection, the vast majority of such endeavors are focused on generating 3D human motion, which is a well-studied research area. Numerous diffusion based methods ranging from body motion \cite{tevet2022human,chen2023executing,azadi2023make,dabral2023mofusion,zhang2024motiondiffuse,shafir2023human} to hand gestures motions \cite{zhang2023diffmotion,baltatzis2023neural}, have emerged in the literature. While most of these methods are text-conditioned, others utilize different conditionings such as 3D landmarks \cite{du2023avatars} and 3D object points \cite{li2023object}. Nevertheless, none of the aforementioned methods tackles the problem of labeled expression animation which is an understudied problem due to the scarcity of 4D facial expression data, unlike speech and human motion data.

    \subsection{Diffusion Models on the 3D Space}
    Although 2D diffusion models have been extensively explored and well-understood, particularly in terms of denoising model architectures and their applications, the exploration in the 3D domain lags behind because of the complexities involved in 3D modeling. Only a handful of studies have explored diffusion processes directly on 3D structures, and these primarily work on point clouds. The pioneering work of \cite{luo2021diffusion} adapted traditional diffusion models for point clouds introducing a probabilistic approach rooted in thermodynamic diffusion processes. Similarly, another approach \cite{zhou20213d} combined diffusion models with point-voxel representations for shape generation. Recent methods have followed more sophisticated approaches by operating on the 3D latent space \cite{vahdat2022lion,lyu2023controllable} taking inspiration from \cite{rombach2022high}. The last two methods can also generate meshes, but they achieve this by reconstructing surfaces from the generated point clouds and lack detail \cite{vahdat2022lion,lyu2023controllable}. To the best of our knowledge, no existing diffusion model directly operates on mesh points while preserving their inherent connectivity to generate meshes.

\section{Method}
\label{sec:method}

    % Add Overview
    Our method learns to animate a mesh of neutral expression towards a target expression, guided by a signal that indicates both the progression and intensity of the resulting expression animation, enabling extensive customization.
    
\begin{figure*}[!h]
        \centering
        \includegraphics[width=0.9\linewidth]{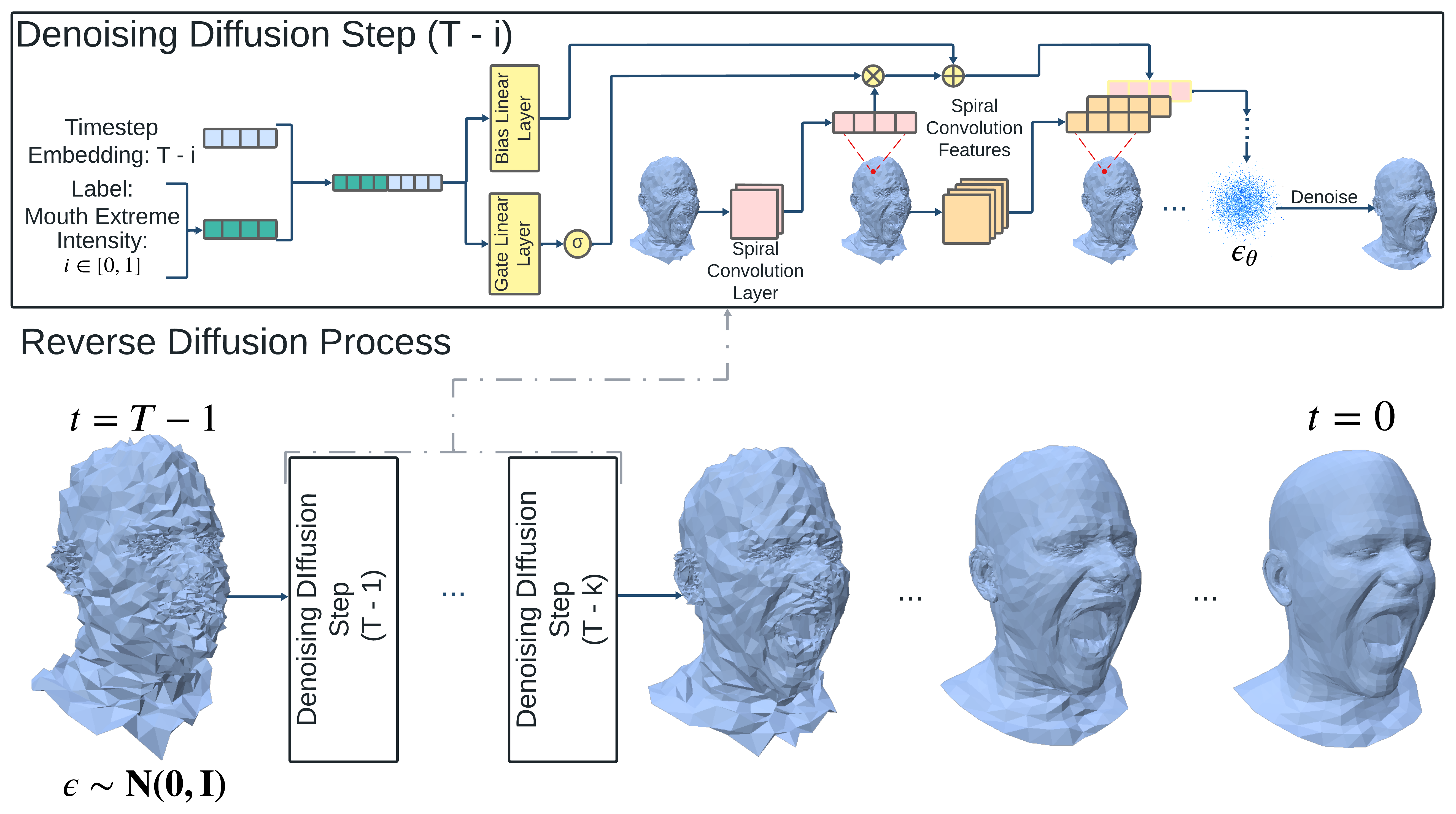}
        \caption{Overview of the proposed frame generation method: Our method generates frames by integrating a point cloud DDPM with an SCN denoising model, conditioned on a concatenated expression and timestep conditioning. It employs spiral convolutional layers, modulating output features with a simple gating and bias mechanism tailored to the conditions. Throughout this process, noise is predicted and systematically subtracted at each timestep until the frame is completely denoised and thus generated. While the method operates on deformations, for visualization, we apply them to the neutral mesh for all timesteps, to show the temporal evolution of the diffusion process.}
        \label{fig:static}
        \vspace{-0.5cm}
    \end{figure*}

\begin{figure*}[!h]
        \centering
        \includegraphics[width=0.9\linewidth]{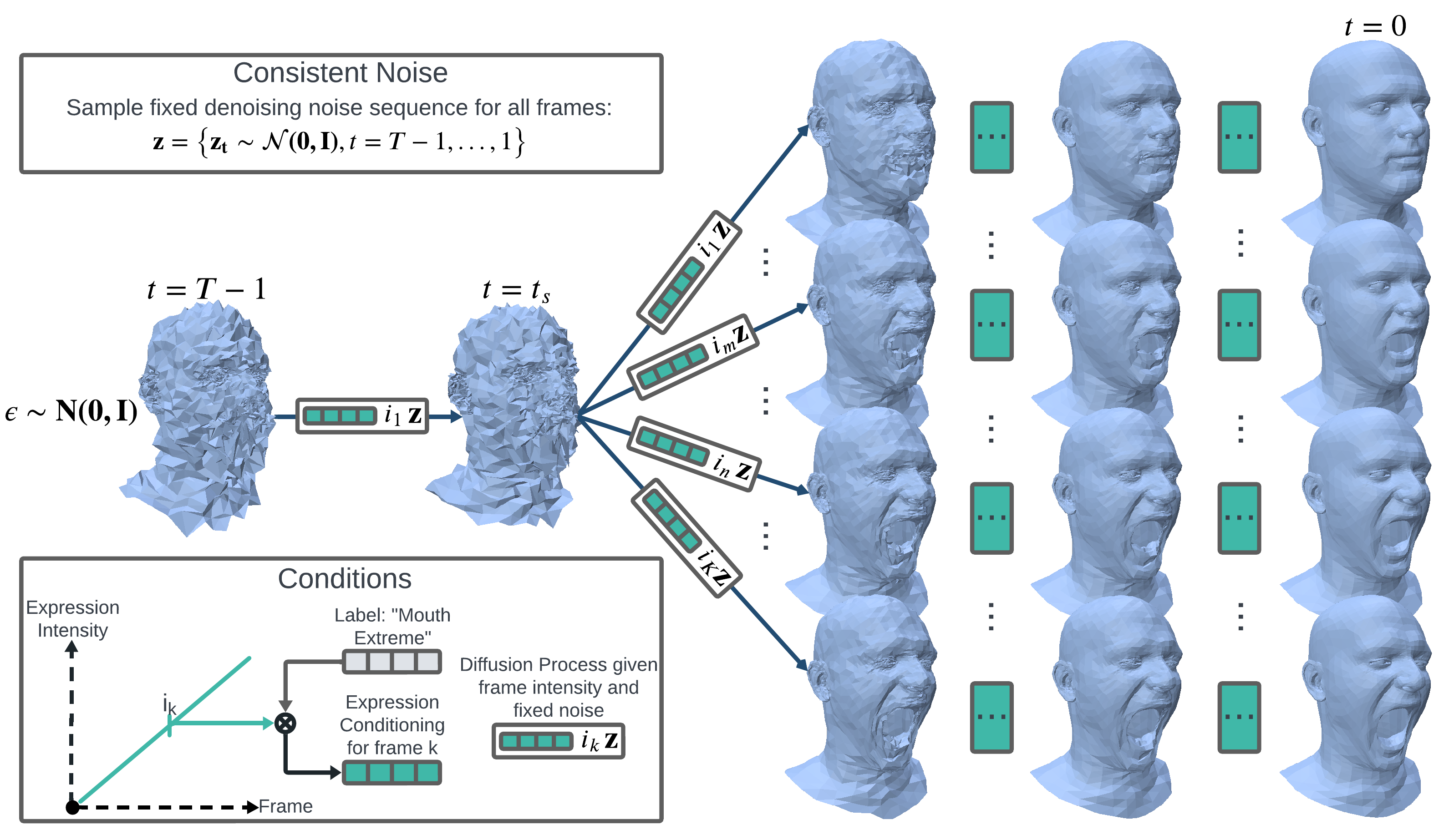}
        \caption{Animation generation via consistent noise sampling: The process initiates by sampling the initial noise $\bf{\epsilon}$ and the denoising noise sequence $\bf{z}$ over $T-1$ timesteps. The diffusion process begins with the first frame using the full range of timesteps. Upon reaching a late denoised stage at $t_s$, the generation for subsequent frames starts in parallel from this denoised state, utilizing only the remaining $t_s$ timesteps. All frames share the same denoising sequence $\bf{z}$, with differences arising from the expression intensity.}
        \label{fig:dynamic}
    \end{figure*}

    To implement the animation mechanism, our framework introduces a novel mesh diffusion process tailored for fixed topology meshes. The frames of each animation used for training are processed by expressing them as deformations from the neutral mesh. Interestingly, while our method model is dynamic, our diffusion model is trained in a conventional static manner. A key feature is our intuitive consistent noise sampling strategy, designed specifically for our problem. This not only ensures temporal coherence, resulting in smooth animations but also accelerates the generation process. \cref{fig:static} provides an overview of our frame generation method, while \cref{fig:dynamic} depicts our consistent noise sampling strategy. In the following subsections, we detail each component and explain the innovations of our method. 

    \subsection{Formulating Diffusion Processes on the Mesh Space}
    \label{sec:method_formulation}
        Our mesh diffusion formulation builds upon \cite{luo2021diffusion}. The core idea of this work is to tailor the diffusion process specifically for point clouds. The objective is to reconstruct a point cloud $\mathbf{X}^{(0)}$ of a desired shape, defined by a shape latent $\mathbf{z} \in \mathbb{R}^{N \times 1}$. This latent is derived from an encoder $\mathbf{z} = E_{\phi}(\mathbf{X}^{(0)})$, and the reverse diffusion process is conditioned on it. Starting from pure noise, the method progresses to generate the point cloud. 

        Consistent with traditional DDPMs, the training objective is to maximize the data's log-likelihood, $\mathbb{E}[\log p_{\bm{\theta}}(\mathbf{X}^{(0)})]$. By leveraging the variational lower bound (ELBO) and further derivations, a simplified training objective is formulated, akin to the approach presented in \cite{ho2020denoising}:

        \begin{equation}
        L(\bm{\theta}, \bm{\phi})=\sum_{i=1}^{N}\left\|\bm{\epsilon} - \bm{\epsilon}_{\theta}(\sqrt{\bar{a}_t} \mathbf{x_i}^{(0)} + \sqrt{1 - \bar{a}_t}\bm{\epsilon}, t, E_{\phi}(\mathbf{X}^{(0)}) )\right\|^2 
        \end{equation}

        Fixed topology meshes are the dominant representation of human faces \cite{egger20203d}. Hence, our objective is to adapt the point cloud diffusion process to operate directly on such meshes. However, challenges arise when examining the original point cloud diffusion process for mesh generation. The primary issue arises from the integration of the loss function, which is focused on denoising at a point level, with the denoising model $\bm{\epsilon}_{\bm{\theta}}(\mathbf{x_i}^{(t)},t,\mathbf{z})$, characterized by its point cloud MLP architecture. The latter treats point clouds as unordered sets due to their inherent permutation invariance. Recognizing that meshes essentially represent strongly structured point clouds defined by their connectivity, such a property becomes a problem, since order and structure are crucial for them. Moreover, point cloud MLPs initially process individual points independently and only aggregate this data in subsequent layers, leading to challenges in accurately capturing the finer local details found in meshes.
        
        To facilitate mesh generation, we employ GNNs as denoising diffusion models in a novel manner. This adaptation leads to a structure-aware, sophisticated diffusion process that not only effectively denoises, but also preserves the connectivity. More specifically, we replace the point cloud MLP denoising model with a Spiral Convolutional Network (SCN) \cite{Bouritsas_2019_ICCV, Gong_2019_ICCV}. This replacement is denoted by $\bm{s_\theta}(\mathbf{x_i}^{(t)},t,\mathbf{c})$ where $\mathbf{c}$ represents the conditioning of the model, serving a role similar to the shape latent $\mathbf{z}$ in the original method.

    \subsection{Training in Line with Static Diffusion Models}
    \label{sec:method_training}
        Typically, dynamic diffusion models that generate modalities, such as videos, train on the entire sequence of animation frames, operating in a whole-animation fashion. In contrast, our approach to training is one frame at a time. Specifically, for a given animation $\mathbf{X} = \{\mathbf{x_i} : i=0,\ldots,K-1\}$, we select a single frame $\mathbf{x_i} \in \mathbb{R}^{N \times 3}$. We then extract the expression stage information for this frame, denoted as $\mathbf{e_i} \in \mathbb{R}^{1 \times M}$, where $M$ represents the number of possible expression categories. This vector $\mathbf{e_i}$ encodes both the category and intensity of the expression, derived from the animation's expression signal $\mathbf{E}=\{\mathbf{e_i} : i=0,\ldots,K-1\}$. Additionally, we obtain the neutral mesh associated with the animation, denoted as $\mathbf{x_0}$. We use it to express the current frame $\mathbf{x_i}$ as deformations relative to it, yielding $\mathbf{d_i} = \mathbf{x_i} - \mathbf{x_0}$ with $\mathbf{d_i} \in \mathbb{R}^{N \times 3}$. From there, training follows the paradigm of static diffusion models with the denoising model conditioned on the expression stage and timestep, using the following loss function:
        \begin{equation}
        L(\bm{\theta})=\sum_{i=1}^{N}\left\|\bm{\epsilon} - \bm{s}_{\theta}(\sqrt{\bar{a}_t} \mathbf{d_i}^{(0)} + \sqrt{1 - \bar{a}_t}\bm{\epsilon}, t, \mathbf{e_i})\right\|^2 
        \end{equation}
        
        A significant advantage of our approach is its computational efficiency. Traditional methods require loading entire animation sequences for training, which is computationally demanding, especially with large meshes. Our frame-by-frame approach avoids this, enabling high-resolution mesh training without performance limitations.

    \subsection{Using the Entire Mesh versus Relying on Landmarks}
    \label{sec:method_entire_mesh}
        We train on the entire mesh, capturing the complex dynamics of facial expressions, including the finest details, drawing inspiration from \cite{potamias2020learning}. This holistic approach, utilizing the comprehensive deformation mesh $\mathbf{d_i} = \mathbf{x_i} - \mathbf{x_0}$ for each frame $i$, surpasses both traditional blendshape-based techniques \cite{cao_faceware,Cheng_2018_CVPR} and landmarks based methods \cite{otberdout2022sparse}. Traditional techniques often fail to represent extreme deformations due to their linear limitations. In contrast, while landmark-based methods capture a significant portion of facial motions through a set of landmarks, they may overlook the fine details of facial dynamics. It is worth noting that even though the methodologies presented in \cite{otberdout2022sparse} yield mesh displacements, their core training still relies on sparse landmarks.

    \subsection{Consistent Noise Sampling}
    \label{sec:method_cons_noise_sampling}
        While the conditioning of the diffusion process on the expression stage might impose some sort of temporal coherence, it is by no means enough for the smoothness required for expression animation. To bridge this gap, building on the ideas presented in \cite{khachatryan2023text2video,wu2023latent,luo2023videofusion}, we propose a modification of the original DDPM sampling algorithm for our problem, named consistent noise sampling, rooted in two primary observations.
        
        Firstly, within the diffusion process, noise drives sample diversity. However, in facial expression animation, maintaining temporal coherence requires careful management of this diversity, as it can otherwise hinder it. To tackle this issue, we propose employing a consistent noise strategy across all animation frames to ensure smooth transitions, acknowledging the minimal differences typically present between consecutive frames. This approach involves applying consistent noise both at the start of denoising and throughout the following denoising steps, ensuring that each frame within an animation maintains coherence. By sampling and maintaining a consistent initial noise implementation $\boldsymbol{\epsilon} \sim \mathcal{N}(\mathbf{0}, \mathbf{I})$ and noise sequence for denoising $\mathbf{z}= {\mathbf{z}_t \sim \mathcal{N}(\mathbf{0}, \mathbf{I}), t=T-1,\ldots,1}$ with $\mathbf{z}_0=\mathbf{0}$ across frames, our model effectively differentiates its output based on the specific expression stage information, ensuring smooth animations.
        
        Secondly, in diffusion models, generation progresses step-wise in a hierarchical manner: earlier timesteps address broader structures, while later steps refine details, such as expression dynamics. Therefore, our method applies the full range of denoising steps $t=T-1,\ldots,0$ for the generation of the first frame of an animation and then reduces steps for the following frames. To implement our approach, we first apply the full range of denoising steps, $t=T-1,\ldots,0$, for generating the initial frame of an animation. Once we reach a late-stage denoised version at timestep $t_s$ for the first frame, denoted as $\mathbf{\hat{d_0}^{t_s}}$, we then initiate the generation of subsequent frames from this advanced denoised state utilizing only the remaining range of timesteps $t = t_s,\ldots,0$. This means that the initial noise prediction for the first frame is given by $\mathbf{\hat{\epsilon_{0}}^{T-1}} = \bm{s_\theta}(\mathbf{\epsilon}, T-1, \mathbf{e_0})$. For subsequent frames, however, the initial noise prediction adjusts to $\mathbf{\hat{\epsilon_i}^{t_s}} = \bm{s_\theta}(\mathbf{\hat{d_0}^{t_s}}, t_s, \mathbf{e_i})$, with $t_s$ effectively serving as the new starting point ($T'-1 = t_s$) for these frames. This strategic adjustment allows for a more efficient generation process by reducing the number of required timesteps while maintaining high fidelity.

        To generate diverse animations for the same expression with our sampling strategy, one simply samples different noise implementations and applies them consistently across all frames. A final advantage of our sampling, which might not be evident at the start, is that once the initial frame is generated, all subsequent frames can be produced concurrently, further improving generation speed.

\section{Experiments}
\label{sec:experiments}
    To demonstrate the superiority of our approach over existing methods, we undertook both quantitative and qualitative experiments focused on 4D facial expression animation using the CoMA \cite{ranjan2018generating} dataset. This dataset features 12 unique identities, each with 12 varied expressions that are extreme and diverse, making it an ideal benchmark for evaluating 4D facial expression generation methods.
    
    Additionally, we applied our approach to the generation of textured 4D facial expressions by training on the MimicMe \cite{papaioannou2022mimicme} dataset, a large-scale database that provides both geometry and textures. We followed the same training protocol used with CoMA for the geometric data and employed a texture animation method using a latent diffusion model. This allowed us to effectively combine the geometry with texture sequences, producing realistic textured 4D facial expressions.
    \subsection{4D Facial Expression Evaluation}
        \subsubsection{Preprocessing}
        \label{sec:experiments_preprocessing}
        
        To ensure meaningful comparisons and address the limitations of \cite{otberdout2022sparse} which is limited to a fixed frame count per animation, all methods are standardized to produce 40 frames. Following the logic of \cite{otberdout2022sparse}, we select subsequences that transition from a neutral to an extreme expression and using interpolation or selection, we ensure consistent length for all. Our preprocessing then quantifies the expression progression by calculating deformations from the neutral mesh for each frame of an animation and smoothing it appropriately. Furthermore, we use global scaling for intensity via an animation extremeness factor, which is necessary for customizable approaches such as ours and \cite{potamias2020learning} to grasp expression intensity information. This factor is normalized across animations of the same expression, providing a consistent framework of assessing expression intensity. Our preprocessing pipeline, designed to equip the model with both progression and intensity information is versatile enough to be applied to any dataset. For more details, we refer the reader to the supplementary material.
    
        \subsubsection{Training Setup}
        \label{sec:experiments_training_setup}
        
        In contrast to \cite{otberdout2022sparse}, we opted to split the CoMA dataset subject-wise to accurately evaluate the generalization of our method. For the experiments, we follow the settings outlined in \cite{otberdout2022sparse} and modify the approach described in \cite{potamias2020learning} to accommodate the CoMA dataset. Our diffusion model's noise schedule uses 1000 steps, starting from $t_1=1e-4$ to $t_T=0.02$. The late denoising strategy is configured with  $t_l=400$, meaning that for subsequent frames beyond the first, we employ only 400 timesteps, initiating from the late denoised version of the first frame, 400 steps before concluding the diffusion process. Our model was trained for 5600 epochs with a batch size of 32, using the Adam optimizer with an initial learning rate of 0.001 and a learning rate scheduler to finally reduce to $1e-4$.

        \subsubsection{Quantitative Evaluation}
        \label{sec:experiments_dataset}

        Given that our training animations transition from neutral to extreme states, we expect \cite{otberdout2022sparse} to inherently learn to generate extreme expressions. To ensure fairness with \cite{otberdout2022sparse}, which isn't designed for customizable intensity levels, we demonstrate the adaptability of our model. We generate expression animations at maximum intensity through our global intensity scaling strategy (Ours-Extreme) and, to mitigate bias, also produce animations across a spectrum of intensities (Ours-Varying), highlighting our method's adaptability. Further enhancing comparison fairness, we apply local intensity scaling (Ours-Local), training our model exclusively on expression progression values. This approach simulates a non-customizable framework, closely mirroring \cite{otberdout2022sparse}, even though it limits the potential of our method. Considering the customizable aspect and significantly worse relative performance of \cite{potamias2020learning}, we pair it with global scaling to achieve the best results, ensuring a truly fair comparison between all methods.
        
        To quantitatively assess the generated expression animations across all methods, we employ the standard metrics. Expression classification is a key benchmark for evaluating 4D facial expression methods. Combining the classifier solutions presented in \cite{otberdout2022sparse,potamias2020learning}, we adopt a similar approach. Our classifier involves a two-stage process. Initially, we use Principal Component Analysis (PCA) to capture the core variations of facial expressions by encoding deformations from the neutral state. This encoding is applied to all animations used to train the models, resulting in a PCA encoder that effectively reduces the dimensionality of the meshes while preserving their expressive spatial features. Following this, we deploy an LSTM-based classifier that operates on these PCA-encoded animations. By processing the temporal sequence of PCA-encoded mesh deformations, our LSTM model effectively captures the dynamic nature of facial expressions. The LSTM output is then sequentially fed through two fully connected layers, with the final layer responsible for class prediction. The second metric utilized for evaluation is the specificity measure, defined as the per-frame average Euclidean distance between the generated animations and the ground truth. This metric serves as an estimate of how closely the generated animations resemble the actual ones, thus acting as a direct indicator of generation quality. When combined with classification accuracy, these metrics together provide a comprehensive framework for the quantitative analysis of 4D facial expressions.
 
        For all subsequent experiments involving our method and MO3DGAN \cite{otberdout2022sparse}, we generate 50 animations per subject and expression due to their stochastic nature. In contrast, for the deterministic LSTM approach \cite{potamias2020learning}, only a single animation is produced for each subject and expression.
        
        \begin{table}[!h]
          \caption{Classification accuracy and average specificity error (mm).}
          \label{tab:method_comparison}
          \centering
          \begin{tabular}{@{}l|c|c@{}}
            \toprule
            Method & Accuracy (\%) & Specificity (mm) \\
            \midrule
            GT & 77.77  & 0 \\
            MO3DGAN \cite{otberdout2022sparse} & 67.22 & 2.18  \\
            LSTM \cite{potamias2020learning} & 45.12 & 3.2  \\
            \hline
                Ours-Extreme & 75.94 & 1.78 \\
                Ours-Local & 72.88  & 1.74 \\ 
                \bf{Ours-Varying} & \bf{78.90} & \bf{1.61} \\
            \bottomrule
          \end{tabular}
        \end{table}

\begin{figure*}[!h]
            \centering
            \includegraphics[width=0.7\linewidth]{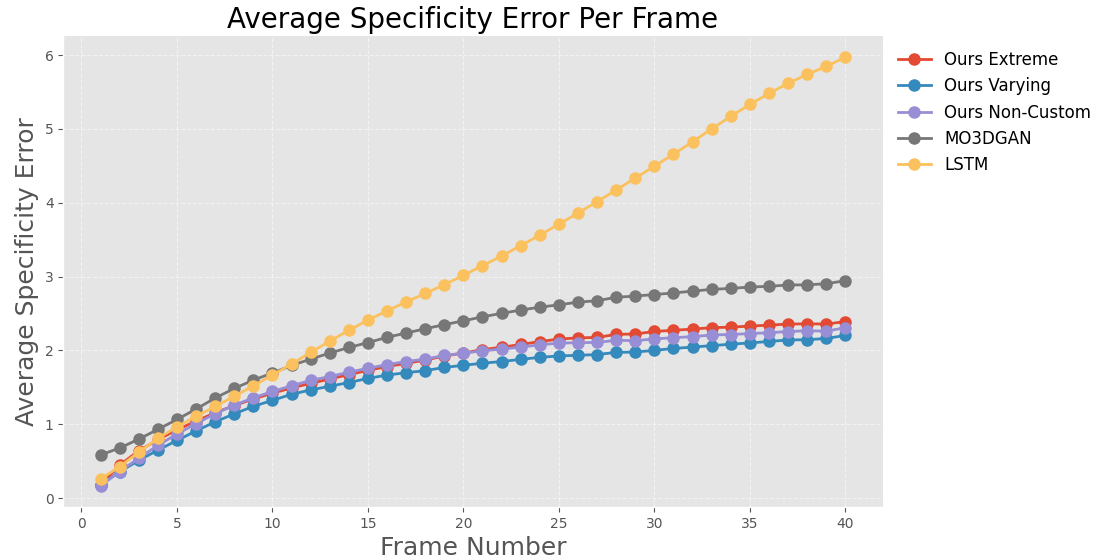}
            \caption{Average per frame specificity error (mm) between the proposed and the baseline methods LSTM \cite{potamias2020learning} and MO3DGAN\cite{otberdout2022sparse}.}
            \label{fig:specificity}
        \end{figure*}

\begin{figure*}[!h]
            \centering
            \includegraphics[width=0.6\linewidth]{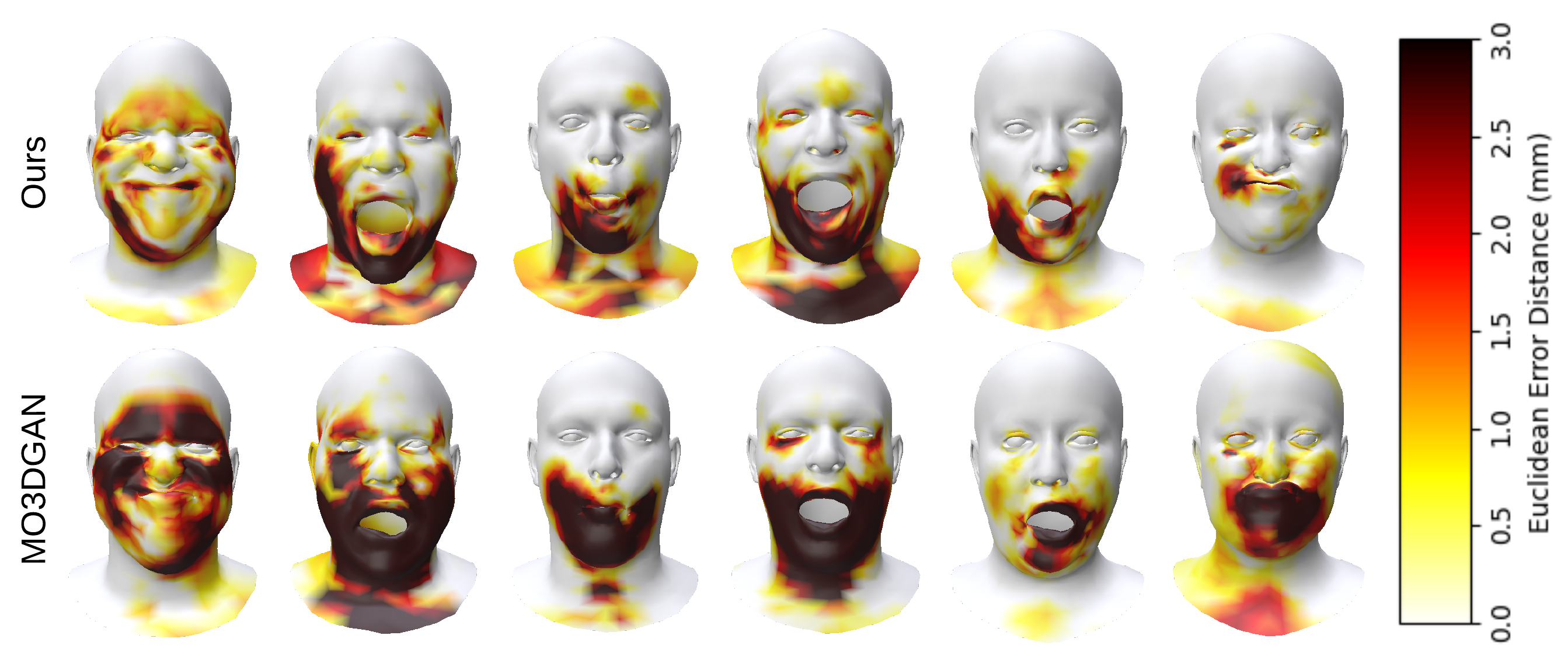}
            \caption{Comparison of final frame generations with their respective ground truths, between ours and MO3DGAN \cite{otberdout2022sparse}. LSTM results are omitted for brevity and due to significantly worse performance.}
            \label{fig:heatmap}
        \end{figure*}

        As can be seen in \cref{tab:method_comparison}, our method surpasses the other two methods across both metrics, demonstrating its superiority. Remarkably, it achieves a similar classification performance with the ground truth, underscoring the quality of our generated expressions. In more detail, the specificity error increases for the final frames, as shown in \cref{fig:specificity}, because these frames correspond to the most extreme expressions. Despite this, every version of our method consistently shows lower error, particularly in these challenging later frames, highlighting our method's capability to accurately generate complex and extreme expressions. To further reinforce our argument about our method's effectiveness in capturing extreme expressions, we include heatmaps in \cref{fig:heatmap}, clearly showcasing our method's preeminence.

        \subsubsection{Qualitative Evaluation}
        \label{sec:experiments_dataset}
        
\begin{figure*}[!h]
            \centering
            \includegraphics[width=\linewidth]{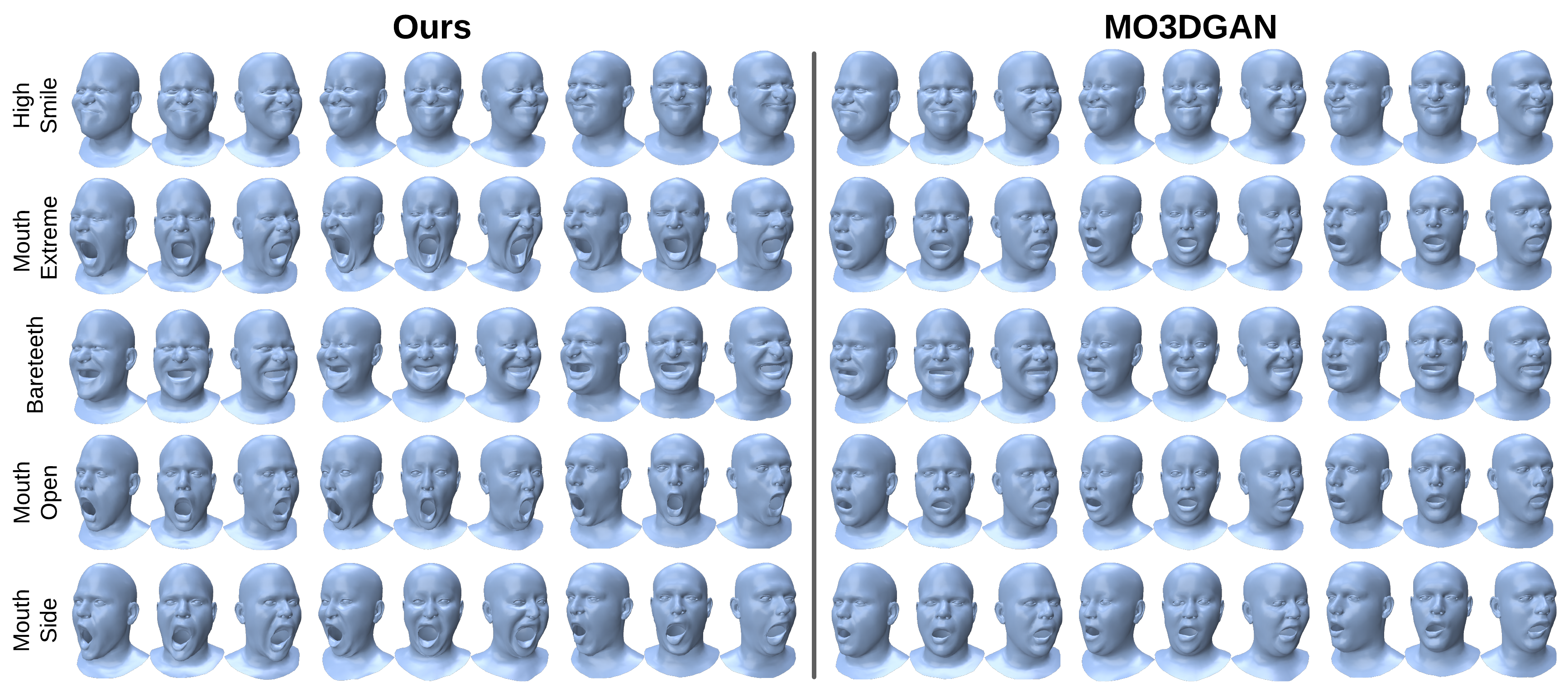}
            \caption{Qualitative comparison of extreme expression generations between ours and MO3DGAN\cite{otberdout2022sparse}. Final expressions are illustrated in the main paper. For full-length dynamic 4D expressions, please refer to the supplementary material.}
            \label{fig:extreme_generations}
        \end{figure*}

\begin{figure*}[!h]
            \centering
            \includegraphics[width=0.75\linewidth]{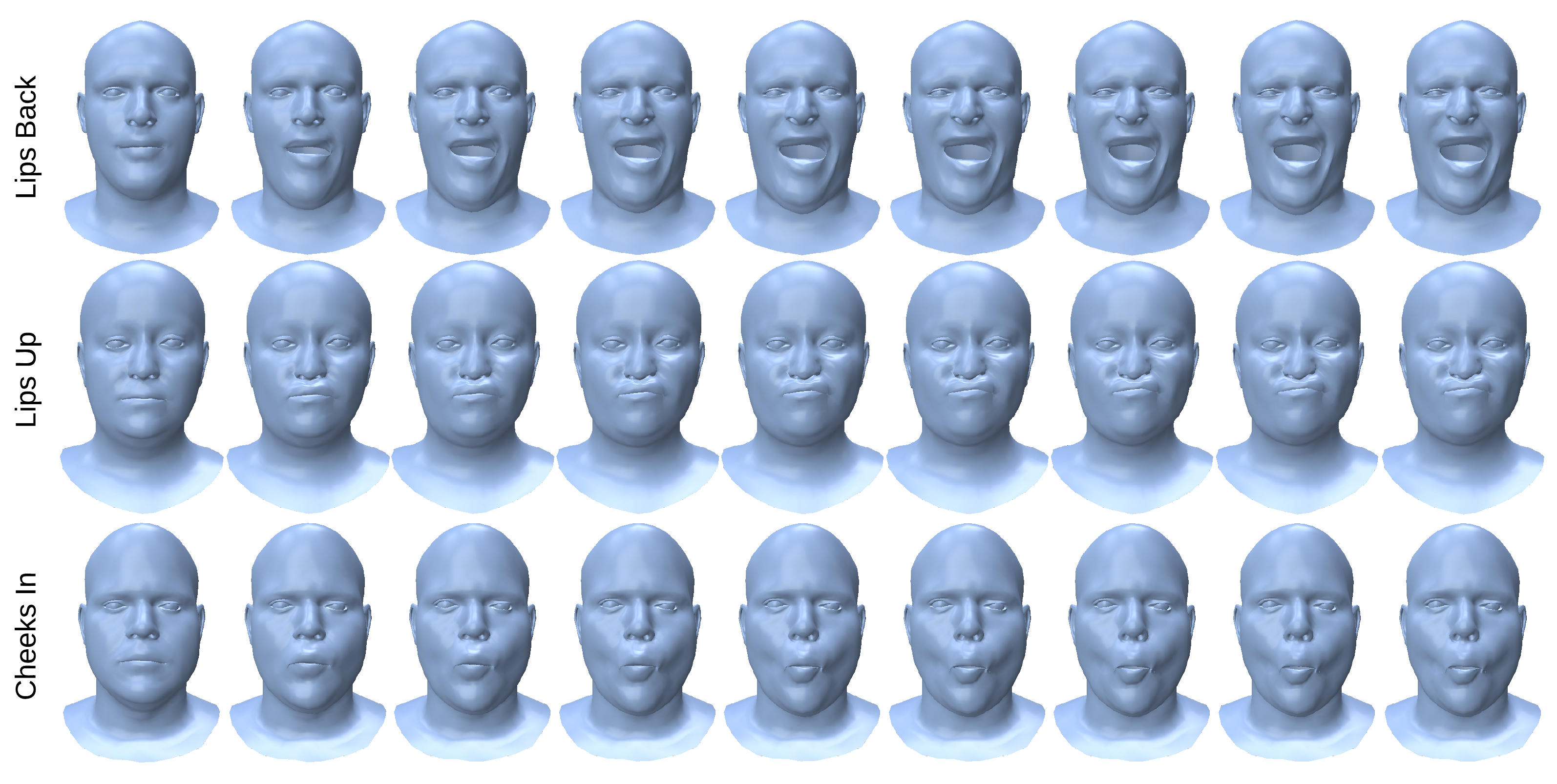}
            \caption{Progression of 4D expressions for three different identities and expressions.}
            \label{fig:expression_progression}
        \end{figure*}

\begin{figure*}[!h]
            \centering
            \includegraphics[width=0.75\linewidth]{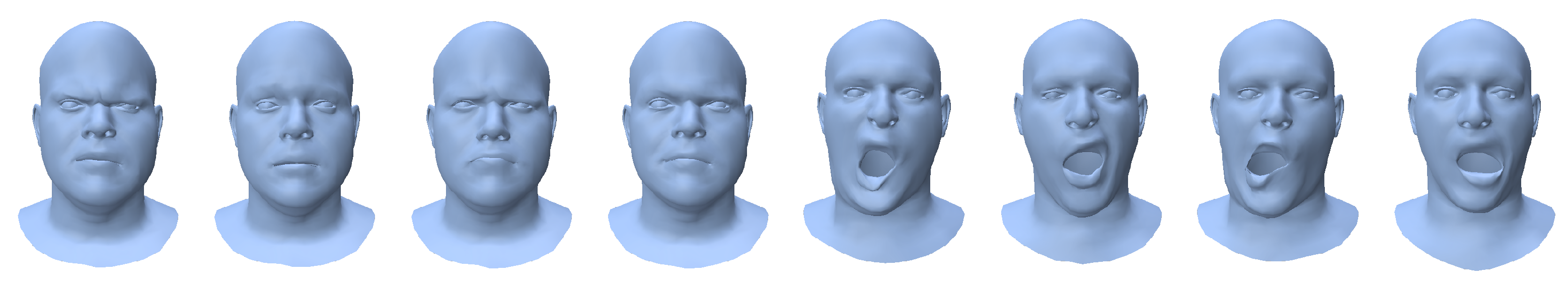}
            \caption{Diversity of generated expressions.}
            \label{fig:diversity}
        \end{figure*}

        Beyond quantitative analysis, we offer qualitative results for a deeper, subjective evaluation. Due to space constraints, we qualitatively compare with the best-performing method. \cref{fig:extreme_generations} showcases generations of extreme expression animations from both methods for a visual comparison. Our method excels by producing more extreme and expressive animations, highlighting our capability to generate high-fidelity expressions. Additionally, as illustrated in \cref{fig:expression_progression}, our method produces animations of high smoothness. Finally, our model excels in creating highly diverse expression animations, with \cref{fig:diversity} showcasing examples from the expression categories with the greatest inherent diversity.

        \subsection{Textured 4D Animation on Large Scale Datasets}
        \label{sec:experiments_dataset}
                
        In our final experiment, we implemented a simple extension of our method to generate textured 4D facial expressions using the diverse MimicMe \cite{papaioannou2022mimicme} dataset, which includes 4,700 subjects, each performing the same expressions. Despite the dataset's variety, it poses challenges such as low frame rates and non-uniform expression initiation points, with some animations transitioning directly between expressions without starting from a neutral state. To overcome these issues, we manually annotated and curated a subset of subjects and their expression animations. By interpolating between frames, we increased the frame count for both texture and geometry. Consequently, we created a refined dataset of 345 subjects, each demonstrating the six fundamental expressions with 40 frames per expression, ensuring a close match between geometry and texture.

        Building on the principles of our geometry diffusion model, we implemented a latent diffusion model (LDM) \cite{rombach2022high} to generate sequences of textures. This approach enhances the standard LDM architecture by conditioning it on the neutral latent (i.e., the neutral texture encoded using the LDM’s encoder), expression intensity, and label of the generated frame. The conditioning mechanisms include channel-wise concatenation for the neutral latent and cross-attention for the expression intensity and label signals. To ensure a meaningful and representative latent space, we trained the autoencoder component of the LDM with all textures from the MimicMe \cite{papaioannou2022mimicme} dataset.
        
        Our training methodology mirrors that of the geometry model, focusing on one frame at a time. This parallel training process ensures a unified learning strategy across both geometry and texture models. For inference, we employ the same consistent noise sampling strategy as used in the geometry model, which has also proven effective for animating textures in our context. Additionally, by utilizing a dataset significantly larger than CoMA's \cite{ranjan2018generating} 12 subjects, we have equipped the geometry diffusion model with identity information. This is achieved by encoding the neutral mesh using a spiral convolutional encoder \cite{Bouritsas_2019_ICCV,Gong_2019_ICCV} that is trained jointly with the diffusion model. The resulting encoding, concatenated with the other conditioning inputs, guides the reverse diffusion process, effectively integrating identity into the generative process.
        
\begin{figure*}[!h]
            \centering
            \includegraphics[width=\linewidth]{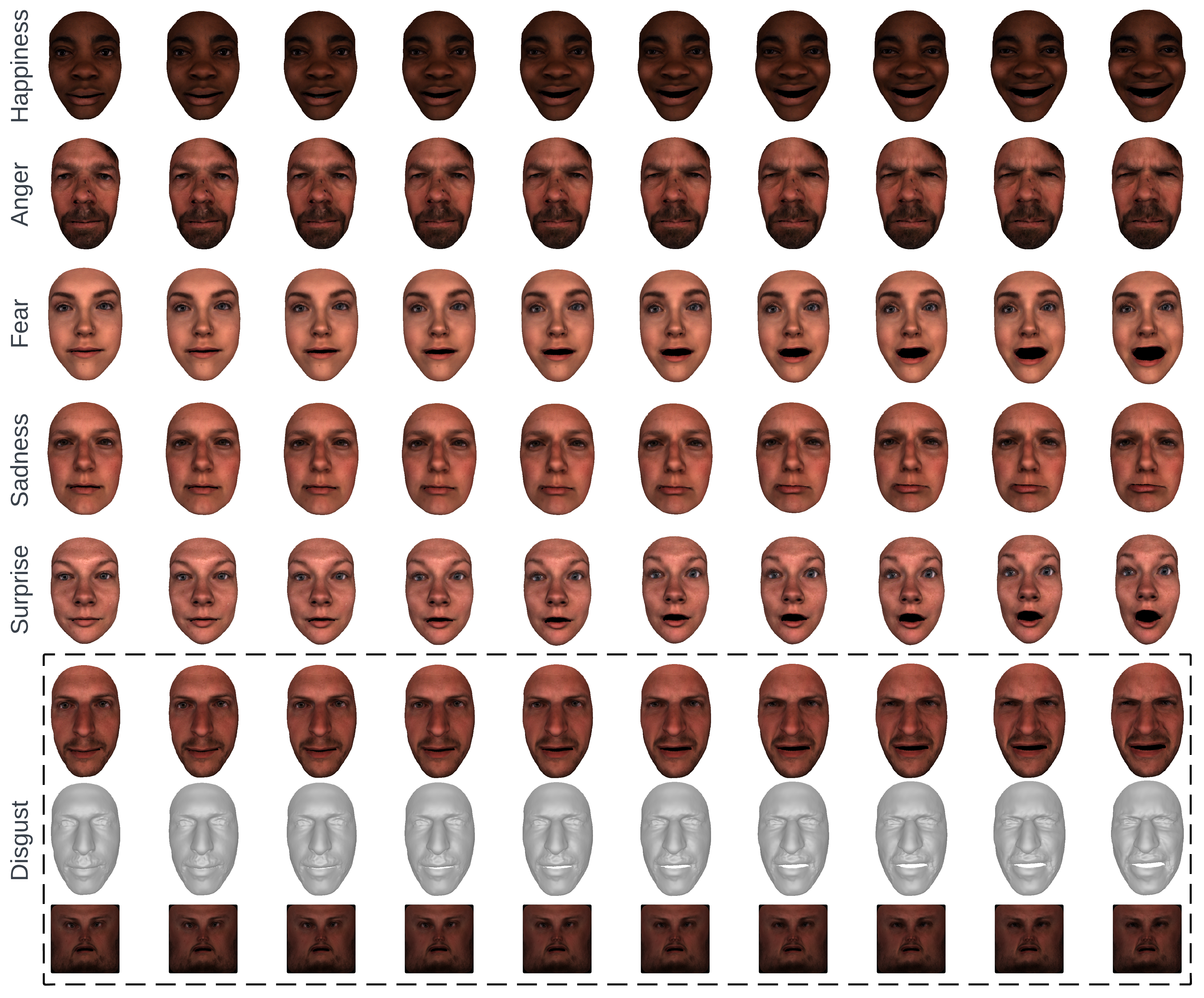}
            \caption{Generated textured 4d facial expressions using our framework. Notably, while our framework doesn’t generate texture on top of geometry, it consistently produces texture sequences that qualitatively align with the corresponding geometries. Expressions progress from left (neutral) to right (apex).}
            \label{fig:mimicme}
        \end{figure*}

        The resulting framework generates textured 4D facial expressions given expression information, a neutral mesh and texture. It achieves this through the application of the geometry model to generate the geometry of the frames and the texture model to create the corresponding textures for each frame. Qualitative results of our method are showcased in \cref{fig:mimicme} for subjective evaluation.
        
\section{Conclusion}
In this work, we introduce AnimateMe, a novel diffusion-based model for fully customizable 4D  expression generation. Leveraging our novel mesh diffusion process with the GNN serving as the denoising model, we facilitate expression generations of high fidelity, significantly surpassing the existing state-of-the-art.
% The qualitative results are also validated quantitatively via extensive evaluation in which we get state-of-the-art results in the standard metrics of classification accuracy and specificity, achieving similar performance to the ground truth in the first metric. 
Paired with our consistent noise sampling strategy, our model ensures the production of smooth animation sequences. We demonstrated the adaptability of our model by extending it to textured animation
% using the same strategy used for the geometry model and provided textured animations for qualitative analysis. 
This extension signifies our method's potential for application in large-scale databases, offering a unified framework for both geometry and texture modeling. To the best of our knowledge, AnimateMe represents the first 4D method to effectively model extreme expressions, addressing a challenge that has not been solved as effectively as possible in the existing literature and the first diffusion model with a GNN as a denoising model.

\clearpage  % TODO REVIEW/FINAL: This \clearpage needs to be removed from both review and camera-ready versions.

% ---- Bibliography ----
%
% BibTeX users should specify bibliography style 'splncs04'.
% References will then be sorted and formatted in the correct style.
%
\bibliographystyle{splncs04}
\bibliography{main}

\begin{thebibliography}{10}
\providecommand{\url}[1]{\texttt{#1}}
\providecommand{\urlprefix}{URL }
\providecommand{\doi}[1]{https://doi.org/#1}

\bibitem{aneja2023facetalk}
Aneja, S., Thies, J., Dai, A., Nießner, M.: Facetalk: Audio-driven motion diffusion for neural parametric head models (2023)

\bibitem{azadi2023make}
Azadi, S., Shah, A., Hayes, T., Parikh, D., Gupta, S.: Make-an-animation: Large-scale text-conditional 3d human motion generation. arXiv preprint arXiv:2305.09662  (2023)

\bibitem{baltatzis2023neural}
Baltatzis, V., Potamias, R.A., Ververas, E., Sun, G., Deng, J., Zafeiriou, S.: Neural sign actors: A diffusion model for 3d sign language production from text. arXiv preprint arXiv:2312.02702  (2023)

\bibitem{blanz2003face}
Blanz, V., Vetter, T.: Face recognition based on fitting a 3d morphable model. IEEE Transactions on pattern analysis and machine intelligence  \textbf{25}(9),  1063--1074 (2003)

\bibitem{blanz2023morphable}
Blanz, V., Vetter, T.: A morphable model for the synthesis of 3d faces. In: Seminal Graphics Papers: Pushing the Boundaries, Volume 2, pp. 157--164 (2023)

\bibitem{blattmann2023align}
Blattmann, A., Rombach, R., Ling, H., Dockhorn, T., Kim, S.W., Fidler, S., Kreis, K.: Align your latents: High-resolution video synthesis with latent diffusion models. In: Proceedings of the IEEE/CVF Conference on Computer Vision and Pattern Recognition. pp. 22563--22575 (2023)

\bibitem{bouritsas2019neural}
Bouritsas, G., Bokhnyak, S., Ploumpis, S., Bronstein, M., Zafeiriou, S.: Neural 3d morphable models: Spiral convolutional networks for 3d shape representation learning and generation. In: Proceedings of the IEEE/CVF International Conference on Computer Vision. pp. 7213--7222 (2019)

\bibitem{Bouritsas_2019_ICCV}
Bouritsas, G., Bokhnyak, S., Ploumpis, S., Bronstein, M., Zafeiriou, S.: Neural 3d morphable models: Spiral convolutional networks for 3d shape representation learning and generation. In: Proceedings of the IEEE/CVF International Conference on Computer Vision (ICCV) (October 2019)

\bibitem{Bouzid_2022}
Bouzid, H., Ballihi, L.: Facial expression video generation based-on spatio-temporal convolutional gan: Fev-gan. Intelligent Systems with Applications  \textbf{16},  200139 (Nov 2022). \doi{10.1016/j.iswa.2022.200139}, \url{http://dx.doi.org/10.1016/j.iswa.2022.200139}

\bibitem{cao_faceware}
Cao, C., Weng, Y., Zhou, S., Tong, Y., Zhou, K.: Facewarehouse: A 3d facial expression database for visual computing. IEEE Transactions on Visualization and Computer Graphics  \textbf{20}(3),  413--425 (2014). \doi{10.1109/TVCG.2013.249}

\bibitem{chen2023executing}
Chen, X., Jiang, B., Liu, W., Huang, Z., Fu, B., Chen, T., Yu, G.: Executing your commands via motion diffusion in latent space. In: Proceedings of the IEEE/CVF Conference on Computer Vision and Pattern Recognition. pp. 18000--18010 (2023)

\bibitem{Cheng_2018_CVPR}
Cheng, S., Kotsia, I., Pantic, M., Zafeiriou, S.: 4dfab: A large scale 4d database for facial expression analysis and biometric applications. In: Proceedings of the IEEE Conference on Computer Vision and Pattern Recognition (CVPR) (June 2018)

\bibitem{cudeiro2019capture}
Cudeiro, D., Bolkart, T., Laidlaw, C., Ranjan, A., Black, M.J.: Capture, learning, and synthesis of 3d speaking styles. In: Proceedings of the IEEE/CVF Conference on Computer Vision and Pattern Recognition. pp. 10101--10111 (2019)

\bibitem{dabral2023mofusion}
Dabral, R., Mughal, M.H., Golyanik, V., Theobalt, C.: Mofusion: A framework for denoising-diffusion-based motion synthesis. In: Proceedings of the IEEE/CVF Conference on Computer Vision and Pattern Recognition. pp. 9760--9770 (2023)

\bibitem{dhariwal2021diffusion}
Dhariwal, P., Nichol, A.: Diffusion models beat gans on image synthesis. Advances in neural information processing systems  \textbf{34},  8780--8794 (2021)

\bibitem{du2023avatars}
Du, Y., Kips, R., Pumarola, A., Starke, S., Thabet, A., Sanakoyeu, A.: Avatars grow legs: Generating smooth human motion from sparse tracking inputs with diffusion model. In: Proceedings of the IEEE/CVF Conference on Computer Vision and Pattern Recognition. pp. 481--490 (2023)

\bibitem{egger20203d}
Egger, B., Smith, W.A., Tewari, A., Wuhrer, S., Zollhoefer, M., Beeler, T., Bernard, F., Bolkart, T., Kortylewski, A., Romdhani, S., et~al.: 3d morphable face models—past, present, and future. ACM Transactions on Graphics (ToG)  \textbf{39}(5),  1--38 (2020)

\bibitem{im2video}
Fan, L., Huang, W., Gan, C., Huang, J., Gong, B.: Controllable image-to-video translation: a case study on facial expression generation. In: Proceedings of the Thirty-Third AAAI Conference on Artificial Intelligence and Thirty-First Innovative Applications of Artificial Intelligence Conference and Ninth AAAI Symposium on Educational Advances in Artificial Intelligence. AAAI'19/IAAI'19/EAAI'19, AAAI Press (2019). \doi{10.1609/aaai.v33i01.33013510}, \url{https://doi.org/10.1609/aaai.v33i01.33013510}

\bibitem{fan2022faceformer}
Fan, Y., Lin, Z., Saito, J., Wang, W., Komura, T.: Faceformer: Speech-driven 3d facial animation with transformers (2022)

\bibitem{Gong_2019_ICCV}
Gong, S., Chen, L., Bronstein, M., Zafeiriou, S.: Spiralnet++: A fast and highly efficient mesh convolution operator. In: Proceedings of the IEEE/CVF International Conference on Computer Vision (ICCV) Workshops (Oct 2019)

\bibitem{he2023latent}
He, Y., Yang, T., Zhang, Y., Shan, Y., Chen, Q.: Latent video diffusion models for high-fidelity long video generation (2023)

\bibitem{ho2022imagen}
Ho, J., Chan, W., Saharia, C., Whang, J., Gao, R., Gritsenko, A., Kingma, D.P., Poole, B., Norouzi, M., Fleet, D.J., Salimans, T.: Imagen video: High definition video generation with diffusion models (2022)

\bibitem{ho2020denoising}
Ho, J., Jain, A., Abbeel, P.: Denoising diffusion probabilistic models. Advances in neural information processing systems  \textbf{33},  6840--6851 (2020)

\bibitem{ho2022video}
Ho, J., Salimans, T., Gritsenko, A., Chan, W., Norouzi, M., Fleet, D.J.: Video diffusion models (2022)

\bibitem{karras2017audio}
Karras, T., Aila, T., Laine, S., Herva, A., Lehtinen, J.: Audio-driven facial animation by joint end-to-end learning of pose and emotion. ACM Transactions on Graphics (TOG)  \textbf{36}(4),  1--12 (2017)

\bibitem{khachatryan2023text2video}
Khachatryan, L., Movsisyan, A., Tadevosyan, V., Henschel, R., Wang, Z., Navasardyan, S., Shi, H.: Text2video-zero: Text-to-image diffusion models are zero-shot video generators. arXiv preprint arXiv:2303.13439  (2023)

\bibitem{li2023object}
Li, J., Wu, J., Liu, C.K.: Object motion guided human motion synthesis. ACM Transactions on Graphics (TOG)  \textbf{42}(6),  1--11 (2023)

\bibitem{luo2021diffusion}
Luo, S., Hu, W.: Diffusion probabilistic models for 3d point cloud generation. In: Proceedings of the IEEE/CVF Conference on Computer Vision and Pattern Recognition. pp. 2837--2845 (2021)

\bibitem{luo2023videofusion}
Luo, Z., Chen, D., Zhang, Y., Huang, Y., Wang, L., Shen, Y., Zhao, D., Zhou, J., Tan, T.: Videofusion: Decomposed diffusion models for high-quality video generation (2023)

\bibitem{lyu2023controllable}
Lyu, Z., Wang, J., An, Y., Zhang, Y., Lin, D., Dai, B.: Controllable mesh generation through sparse latent point diffusion models. In: Proceedings of the IEEE/CVF Conference on Computer Vision and Pattern Recognition. pp. 271--280 (2023)

\bibitem{ma2024diffspeaker}
Ma, Z., Zhu, X., Qi, G., Qian, C., Zhang, Z., Lei, Z.: Diffspeaker: Speech-driven 3d facial animation with diffusion transformer. arXiv preprint arXiv:2402.05712  (2024)

\bibitem{wgan2d}
Otberdout, N., Daoudi, M., Kacem, A., Ballihi, L., Berretti, S.: Dynamic facial expression generation on hilbert hypersphere with conditional wasserstein generative adversarial nets. IEEE Transactions on Pattern Analysis and Machine Intelligence  \textbf{44}(2),  848--863 (2022). \doi{10.1109/TPAMI.2020.3002500}

\bibitem{otberdout2022sparse}
Otberdout, N., Ferrari, C., Daoudi, M., Berretti, S., Bimbo, A.D.: Sparse to dense dynamic 3d facial expression generation (2022)

\bibitem{papaioannou2022mimicme}
Papaioannou, A., Gecer, B., Cheng, S., Chrysos, G., Deng, J., Fotiadou, E., Kampouris, C., Kollias, D., Moschoglou, S., Songsri-In, K., et~al.: Mimicme: A large scale diverse 4d database for facial expression analysis. In: European Conference on Computer Vision. pp. 467--484 (2022)

\bibitem{park2023said}
Park, I., Cho, J.: Said: Speech-driven blendshape facial animation with diffusion. arXiv preprint arXiv:2401.08655  (2023)

\bibitem{peng2023emotalk}
Peng, Z., Wu, H., Song, Z., Xu, H., Zhu, X., He, J., Liu, H., Fan, Z.: Emotalk: Speech-driven emotional disentanglement for 3d face animation. In: Proceedings of the IEEE/CVF International Conference on Computer Vision. pp. 20687--20697 (2023)

\bibitem{pham2017speech}
Pham, H.X., Cheung, S., Pavlovic, V.: Speech-driven 3d facial animation with implicit emotional awareness: A deep learning approach. In: Proceedings of the IEEE conference on computer vision and pattern recognition workshops. pp. 80--88 (2017)

\bibitem{ploumpis2020towards}
Ploumpis, S., Ververas, E., O'Sullivan, E., Moschoglou, S., Wang, H., Pears, N., Smith, W.A., Gecer, B., Zafeiriou, S.: Towards a complete 3d morphable model of the human head. IEEE transactions on pattern analysis and machine intelligence  \textbf{43}(11),  4142--4160 (2020)

\bibitem{ploumpis2019combining}
Ploumpis, S., Wang, H., Pears, N., Smith, W.A., Zafeiriou, S.: Combining 3d morphable models: A large scale face-and-head model. In: Proceedings of the IEEE/CVF Conference on Computer Vision and Pattern Recognition. pp. 10934--10943 (2019)

\bibitem{potamias2020learning}
Potamias, R.A., Zheng, J., Ploumpis, S., Bouritsas, G., Ververas, E., Zafeiriou, S.: Learning to generate customized dynamic 3d facial expressions (2020)

\bibitem{ranjan2018generating}
Ranjan, A., Bolkart, T., Sanyal, S., Black, M.J.: Generating 3d faces using convolutional mesh autoencoders. In: Proceedings of the European conference on computer vision (ECCV). pp. 704--720 (2018)

\bibitem{Richard_2021_ICCV}
Richard, A., Zollh\"ofer, M., Wen, Y., de~la Torre, F., Sheikh, Y.: Meshtalk: 3d face animation from speech using cross-modality disentanglement. In: Proceedings of the IEEE/CVF International Conference on Computer Vision (ICCV). pp. 1173--1182 (October 2021)

\bibitem{rombach2022high}
Rombach, R., Blattmann, A., Lorenz, D., Esser, P., Ommer, B.: High-resolution image synthesis with latent diffusion models. In: Proceedings of the IEEE/CVF conference on computer vision and pattern recognition. pp. 10684--10695 (2022)

\bibitem{shafir2023human}
Shafir, Y., Tevet, G., Kapon, R., Bermano, A.H.: Human motion diffusion as a generative prior. arXiv preprint arXiv:2303.01418  (2023)

\bibitem{singer2023makeavideo}
Singer, U., Polyak, A., Hayes, T., Yin, X., An, J., Zhang, S., Hu, Q., Yang, H., Ashual, O., Gafni, O., Parikh, D., Gupta, S., Taigman, Y.: Make-a-video: Text-to-video generation without text-video data. In: The Eleventh International Conference on Learning Representations (2023), \url{https://openreview.net/forum?id=nJfylDvgzlq}

\bibitem{song2022denoising}
Song, J., Meng, C., Ermon, S.: Denoising diffusion implicit models (2022)

\bibitem{song2020score}
Song, Y., Sohl-Dickstein, J., Kingma, D.P., Kumar, A., Ermon, S., Poole, B.: Score-based generative modeling through stochastic differential equations. arXiv preprint arXiv:2011.13456  (2020)

\bibitem{stan2023facediffuser}
Stan, S., Haque, K.I., Yumak, Z.: Facediffuser: Speech-driven 3d facial animation synthesis using diffusion. In: Proceedings of the 16th ACM SIGGRAPH Conference on Motion, Interaction and Games. pp. 1--11 (2023)

\bibitem{tevet2022human}
Tevet, G., Raab, S., Gordon, B., Shafir, Y., Cohen-Or, D., Bermano, A.H.: Human motion diffusion model. arXiv preprint arXiv:2209.14916  (2022)

\bibitem{thambiraja20233diface}
Thambiraja, B., Aliakbarian, S., Cosker, D., Thies, J.: 3diface: Diffusion-based speech-driven 3d facial animation and editing. arXiv preprint arXiv:2312.00870  (2023)

\bibitem{Thambiraja_2023_ICCV}
Thambiraja, B., Habibie, I., Aliakbarian, S., Cosker, D., Theobalt, C., Thies, J.: Imitator: Personalized speech-driven 3d facial animation. In: Proceedings of the IEEE/CVF International Conference on Computer Vision (ICCV). pp. 20621--20631 (October 2023)

\bibitem{tran2018nonlinear}
Tran, L., Liu, X.: Nonlinear 3d face morphable model. In: Proceedings of the IEEE conference on computer vision and pattern recognition. pp. 7346--7355 (2018)

\bibitem{tulyakov2017mocogan}
Tulyakov, S., Liu, M.Y., Yang, X., Kautz, J.: Mocogan: Decomposing motion and content for video generation (2017)

\bibitem{tzirakis2020synthesising}
Tzirakis, P., Papaioannou, A., Lattas, A., Tarasiou, M., Schuller, B., Zafeiriou, S.: Synthesising 3d facial motion from “in-the-wild” speech. In: 2020 15th IEEE International Conference on Automatic Face and Gesture Recognition (FG 2020). pp. 265--272 (2020)

\bibitem{vahdat2022lion}
Vahdat, A., Williams, F., Gojcic, Z., Litany, O., Fidler, S., Kreis, K., et~al.: Lion: Latent point diffusion models for 3d shape generation. Advances in Neural Information Processing Systems  \textbf{35},  10021--10039 (2022)

\bibitem{g3an}
Wang, Y., Bilinski, P., Bremond, F., Dantcheva, A.: G3an: Disentangling appearance and motion for video generation. In: 2020 IEEE/CVF Conference on Computer Vision and Pattern Recognition (CVPR). pp. 5263--5272 (2020). \doi{10.1109/CVPR42600.2020.00531}

\bibitem{wu2023latent}
Wu, C.H., De~la Torre, F.: A latent space of stochastic diffusion models for zero-shot image editing and guidance. In: Proceedings of the IEEE/CVF International Conference on Computer Vision. pp. 7378--7387 (2023)

\bibitem{f3agan}
Wu, X., Zhang, Q., Wu, Y., Wang, H., Li, S., Sun, L., Li, X.: F3a-gan: Facial flow for face animation with generative adversarial networks (05 2022)

\bibitem{xing2023codetalker}
Xing, J., Xia, M., Zhang, Y., Cun, X., Wang, J., Wong, T.T.: Codetalker: Speech-driven 3d facial animation with discrete motion prior. In: Proceedings of the IEEE/CVF Conference on Computer Vision and Pattern Recognition. pp. 12780--12790 (2023)

\bibitem{zhang2023diffmotion}
Zhang, F., Ji, N., Gao, F., Li, Y.: Diffmotion: Speech-driven gesture synthesis using denoising diffusion model. In: International Conference on Multimedia Modeling. pp. 231--242. Springer (2023)

\bibitem{zhang2024motiondiffuse}
Zhang, M., Cai, Z., Pan, L., Hong, F., Guo, X., Yang, L., Liu, Z.: Motiondiffuse: Text-driven human motion generation with diffusion model. IEEE Transactions on Pattern Analysis and Machine Intelligence  (2024)

\bibitem{zhou2023magicvideo}
Zhou, D., Wang, W., Yan, H., Lv, W., Zhu, Y., Feng, J.: Magicvideo: Efficient video generation with latent diffusion models (2023)

\bibitem{zhou20213d}
Zhou, L., Du, Y., Wu, J.: 3d shape generation and completion through point-voxel diffusion. In: Proceedings of the IEEE/CVF International Conference on Computer Vision. pp. 5826--5835 (2021)

\end{thebibliography}

\clearpage

% Manually create a heading for the supplementary material
\begin{center}
    % Supplementary Material Title
    \vspace*{5pt} % Adjust space as needed
    {\Large \textbf{AnimateMe: 4D Facial Expressions via Diffusion Models \\ (Supplementary Material)}\par}
\end{center}

% Reset section counter and modify section numbering for supplementary material
\setcounter{section}{0} % Resets the section counter
\renewcommand{\thesection}{\Alph{section}} % Changes section numbering to alphabetical

\section{Additional Implementation Details}
\label{sec:implement}
    \subsection{Preprocessing Details}
    To standardize animations to a uniform 40 frames, we employ a combination of selection and interpolation techniques. For sequences exceeding 40 frames, we select the most distinct frames in terms of their differences from consecutive frames. On the other hand, for sequences with fewer than 40 frames, we employ interpolation between the frames that exhibit the greatest differences, thereby seamlessly expanding to the required frame count.
    
    Regarding the extremeness factors, they are derived by calculating deformations at key facial landmarks from neutral to apex frames for each animation. These factors are then normalized against the maximum extremeness factor within the same expression category, as mentioned in the main paper.

    \subsection{Dataset Splitting Selection}
    Regarding dataset splitting, our approach diverges from that of \cite{otberdout2022sparse}. Whereas \cite{otberdout2022sparse} utilizes the first subsequence from each subject and expression from their divided subsequences for training and allocate the rest for testing, we adopt the split methodology from the subject-independent CoMA \cite{ranjan2018generating} experiments. This involves training on nine identities and testing on three, encompassing all their expression animations. This decision was made to ensure the model remains unexposed to the specific landmarks or deformation patterns of the test identities during training, thus avoiding potential overfitting. Ideally, the models should be completely unfamiliar with the test data to ensure fair evaluation conditions. Therefore, our choice aligns with the goals of evaluating model generalizability, offering a more accurate measure of its performance on unseen data.

\section{Consistent Noise Sampling Algorithm}
\label{sec:cons_noise}
For the sake of completeness, we also provide the analytical algorithm for consistent noise sampling, ensuring that the notation remains consistent with that used in the main paper.
\begin{algorithm}[H]
\caption{Consistent Noise Sampling}
\begin{algorithmic}[1]
\small
\State Get the neutral mesh to be animated: $\mathbf{x_0} \in \mathbb{R}^{N \times 3}$
\State Get the expression progression signal: $\mathbf{E}=\{\mathbf{e_i} : i=0,\ldots,K-1\}$
\State
\State $\boldsymbol{\epsilon} \sim \mathcal{N}(\mathbf{0},\mathbf{I})$
\State ${\mathbf{z}_t \sim \mathcal{N}(\mathbf{0}, \mathbf{I}), t=T,\ldots,2}$
\State $\mathbf{z}_1=\mathbf{0}$
\State
\State Sample the expression progression for the first frame: $\mathbf{e_0}$
\State $\mathbf{\hat{d_0}^{T}} = \boldsymbol{\epsilon}$
\For{$t = T$ to $t_s + 1$}
    \State \(\mathbf{\hat{d_{0}}^{t-1}} = \frac{1}{\sqrt{a_t}} \left (\mathbf{\hat{d_{0}}^{t}} - \frac{1-a_t}{\sqrt{1-\bar{a}_t}}\mathbf{s_{\theta}}(\mathbf{\hat{d_{0}}^{t}}, t,\mathbf{e_0}) \right) + \sigma_t \mathbf{z_t}\)
\EndFor
\State
\State $\mathbf{\hat{d_{s}}} = \mathbf{\hat{d_{0}}^{t_s}}$

% Here is the line modified to include "in parallel"
\For{$k = 0$ to $K-1$ \textbf{in parallel}}
    \State Sample the expression progression for the current frame: $\mathbf{e_i}$
    \State $\mathbf{\hat{d_{k}}^{t_s}} = \mathbf{\hat{d_{s}}}$
    \For{$t = t_s$ to $1$}
        \State \(\mathbf{\hat{d_{k}}^{t-1}} = \frac{1}{\sqrt{a_t}} \left (\mathbf{\hat{d_{k}}^{t}} - \frac{1-a_t}{\sqrt{1-\bar{a}_t}}\mathbf{s_{\theta}}(\mathbf{\hat{d_{k}}^{t}}, t,\mathbf{e_k}) \right) + \sigma_t \mathbf{z_t}\)
    \EndFor
    \State Apply the deformations to the neutral mesh: $\mathbf{\hat{x_k}} = \mathbf{x_0} + \mathbf{\hat{d_{k}}^{0}}$
\EndFor
\end{algorithmic}
\end{algorithm}

\section{Architectural Details}
Besides transitioning from an MLP to an SCN for the denoising model architecture, additional adjustments have been implemented to optimize mesh generation. The model is enhanced through the integration of learnable index embeddings with a small dimensionality, which are added to the point clouds as additional features for each point. These embeddings not only improve the smoothness of the generated point clouds but also assist with the maintenance of the mesh's connectivity. Further refinement is achieved with the use of learnable timestep embeddings of increased dimensions. This adaptation allows for the capture of complex patterns essential for the success of the diffusion process. Additionally, the architecture is designed to prioritize the preservation of details and the recognition of complex patterns in 3D data. It features an increase in both the number of filters and the sequence length progressively deeper into the network. This strategy ensures a hierarchical representation of the data, enabling the capture of increasingly abstract and complex features. Finally, to maintain the complexity and detail of the 3D mesh data across all layers, the architecture avoids downsampling. This decision ensures that no vital information is lost throughout the denoising process.

\section{Additional Qualitative Results}
We provide additional qualitative results to showcase the diversity of the generated expressions of our model. Please note that we only provide the last frame of the expression animations.

    \begin{figure*}[!h]
        \centering
        \includegraphics[width=\linewidth]{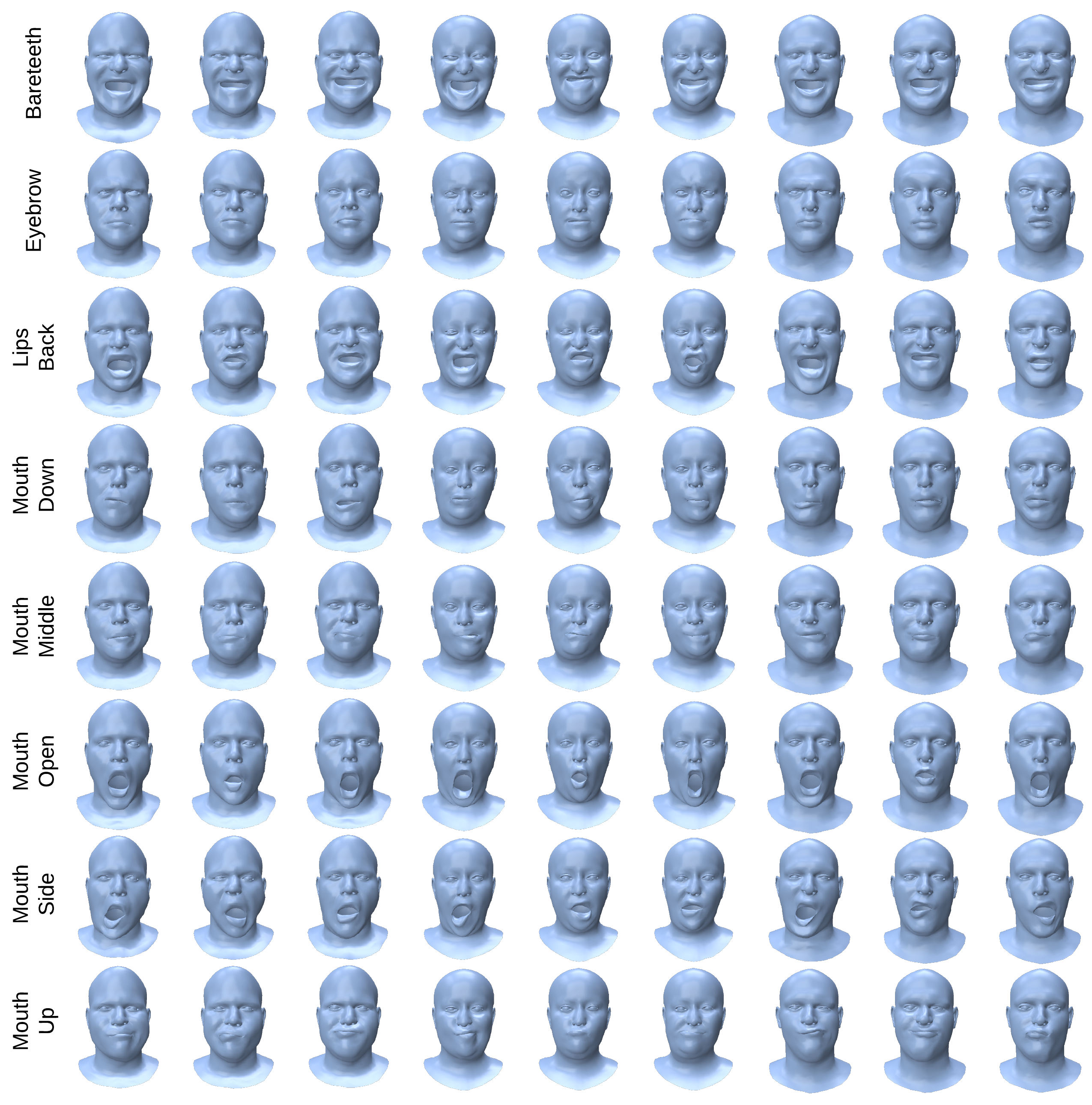}
        \caption{Diversity of generated expressions.}
        \label{fig:diversity_more}
    \end{figure*}

\section{Method Limitations}
We acknowledge that while our method represents a significant advancement in the generation of 4D facial expressions, surpassing previous works, it is not without its limitations. Firstly, our approach is inherently conditioned on an expression progression signal. This conditioning, despite enabling customizable generation, is not ideal due to its restrictive nature. Future iterations of our work could benefit from exploring additional conditioning options to enhance versatility. Secondly, our method's reliance on mesh representations, stemming from the use of a Graph Neural Network (GNN) based denoising model, limits its applicability to other 3D representation forms. This specialization, while effective, narrows the scope of our method's utility. Thirdly, the adoption of a diffusion approach inherently slows down our method, a drawback that becomes more pronounced with larger meshes. Nevertheless, in our textured experiment, downsampling high-resolution meshes (nearly 28K vertices) prior to processing and then upsampling the results has proven effective in mitigating speed issues. Future work could explore integrating DDIM \cite{song2022denoising} sampling with our diffusion model to accelerate generation. Additionally, while not a primary concern, the generation speed of our texture LDM \cite{rombach2022high} also presents challenges. However, by employing DDIM sampling with just 200 timesteps, we achieved very good results. Looking ahead, conditioning the LDM on the geometry of each frame could significantly enhance the coherence between generated textures and geometries, marking a direction for future research.

\section{Ethical Limitations}
Our method animates static facial meshes to match specific expressions, opening up new possibilities in digital media. However, it also presents ethical challenges, particularly concerning misuse and consent. The ease with which our method can animate faces raises concerns about its potential use in creating deepfakes. These manipulated 3D animations can mislead people or harm someone's reputation without their permission. Equally important is the issue of consent and ownership. Using someone's likeness to animate expressions without their clear approval crosses ethical boundaries, especially if those animations are used in ways the person wouldn't agree with. These ethical considerations highlight the need for clear guidelines and consent protocols, ensuring that our technology is used responsibly and respects individual privacy and rights.

\end{document}